\documentclass[10pt,twocolumn,letterpaper]{article}

\usepackage{color}
\usepackage{caption}
\usepackage[pagenumbers]{iccv} %

\definecolor{iccvblue}{rgb}{0.21,0.49,0.74}
\usepackage[pagebackref,breaklinks,colorlinks,allcolors=iccvblue]{hyperref}

\title{ChatAnyone: Stylized Real-time Portrait Video Generation with Hierarchical Motion Diffusion Model}

\author{
    Jinwei Qi \quad Chaonan Ji \quad Sheng Xu \quad Peng Zhang \quad Bang Zhang \quad Liefeng Bo \\
    Tongyi Lab, Alibaba Group  \\
    \tt\small \url{https://humanaigc.github.io/chat-anyone/}
}

\begin{document}
\twocolumn[{
    \renewcommand\twocolumn[1][]{#1}
    \maketitle
    \vspace*{-2.9em}
    \begin{center}
        \captionsetup{type=figure}
        \includegraphics[width=1.0\linewidth]{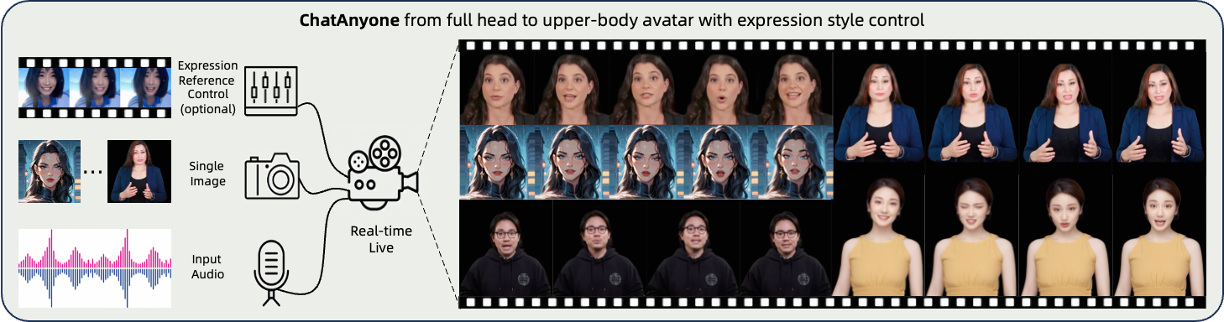} \vspace{-0.8em}
        \captionof{figure}{
            \textbf{Illustration of real-time portrait video generation.} Given a portrait image and audio sequence as input, our model can generate high-fidelity animation results from full head to upper-body interaction with diverse facial expressions and style control.
        }
        \label{fig:overview}
    \end{center}
}]

\begin{abstract}
Real-time interactive video-chat portraits have been increasingly recognized as the future trend, particularly due to the remarkable progress made in text and voice chat technologies. 
However, existing methods primarily focus on real-time generation of head movements, but struggle to produce synchronized body motions that match these head actions. Additionally, achieving fine-grained control over the speaking style and nuances of facial expressions remains a challenge. 
To address these limitations, we introduce a novel framework for stylized real-time portrait video generation, enabling expressive and flexible video chat that extends from talking head to upper-body interaction. Our approach consists of the following two stages. 
The first stage involves efficient hierarchical motion diffusion models, that take both explicit and implicit motion representations into account based on audio inputs, which can generate a diverse range of facial expressions with stylistic control and synchronization between head and body movements. 
The second stage aims to generate portrait video featuring upper-body movements, including hand gestures. We inject explicit hand control signals into the generator to produce more detailed hand movements, and further perform face refinement to enhance the overall realism and expressiveness of the portrait video. 
Additionally, our approach supports efficient and continuous generation of upper-body portrait video in maximum $512 \times 768$ resolution at up to 30fps on 4090 GPU, supporting interactive video-chat in real-time.
Experimental results demonstrate the capability of our approach to produce portrait videos with rich expressiveness and natural upper-body movements.
\end{abstract}

\section{Introduction}
\label{sec:intro}

In recent years, the rapid advancement of LLMs and diffusion models has significantly enhanced the capabilities of text and voice chat \cite{chu2023qwenaudio,moshi}, which achieves astonishing results in interactive AI conversation. Looking ahead, real-time interactive digital humans in video formats represent the future trend, promising even more immersive and truly lifelike interactions. Consequently, as shown in Figure \ref{fig:overview}, there is a growing demand for highly expressive portrait videos that not only exhibit natural head poses and facial expressions, but also fluid and contextually appropriate body movements, creating an authentic and engaging user experience. 

Achieving highly expressive real-time interactive portrait video generation has always been a hot research topic. Early works \cite{wav2lip,speech2lip} leverage the advancements in Generative Adversarial Networks (GANs) \cite{gan} to focus on lip sync, predicting lip motion in mouth region. Further research \cite{sadtalker} extends the scope to generate overall head movements from a single image, including not only facial expressions but also head poses. Additionally, some studies go a step forward to introduce basic emotion control \cite{emotalk} or expand the approach to generate simple body movements \cite{latentCospeech}. Despite these advancements, there is still a long way to go for achieving fine-grained emotionally controllable and photo-realistic portrait video generation.

With the rise of diffusion models recently, many efforts have explored the powerful generative capabilities of diffusion models to achieve higher expressiveness in both face and body dynamics \cite{emo, emo2, vlogger}. However, these methods often struggle with real-time inference due to their computational complexity and the iterative nature of denoising process, which makes them fail to meet the latency requirements for real-time interactions. 
In addition to the aforementioned approaches, some recent works \cite{vasa, ifmdm} design lightweight diffusion models, which learn audio to facial motion representation mappings, by taking advantage of diverse generative capabilities of diffusion models for expressive face motion generation. Subsequently, they employ efficient GANs to generate high-quality head images. 

Although these methods make a step forward to balance the trade-off between efficiency and diversity in talking head contents, these methods still fall short in fine-grained expression control, for example, achieving subtle variations in expression intensity or replicating specific emotional style to particular individuals. Besides, generating highly realistic textures and detailed hand gestures also remains challenging.

To address these limitations and achieve more realistic real-time portrait video interactions, we need to focus on three key areas: 
1) \textit{Fine-grained expression control and style transfer}: Achieve precise control over facial expression and head pose, while enabling style transfer to enhance expressiveness and personalization.
2) \textit{Coordinated body dynamics including hand gestures}: Generate natural body movements including detailed hand gestures, which also need to be synchronized with facial expressions for realistic interaction.
3) \textit{High extensibility and real-time inference efficiency}: Support extensible real-time generation from head-only to upper-body animations, ensuring high performance and efficiency for interaction applications. 

In response to the above challenges, we introduce a novel framework for stylized real-time portrait video generation including two stages, namely upper-body motion representation generation from audio inputs and high-quality portrait video generation from a single image in real-time. Our approach enables expressive and flexible video chat which can extend from talking head to upper-body interaction, significantly enhancing the realism and engagement of digital human interactions. The key contributions of our work are as follows:
\begin{itemize}
    \item \textbf{Efficient Hierarchical motion diffusion model} is proposed for the first stage to generate face and body control signals hierarchically based on input audio, considering both explicit and implicit motion signals for precise facial expressions. Furthermore, fine-grained expression control is introduced to realize different variations in expression intensity, as well as stylistic expression transfer from reference videos, which aims to produce controllable and personalized expressions.
    \item \textbf{Hybrid control fusion generative model} is designed for the second stage, which utilizes explicit landmarks for direct and editable facial expression generation, while implicit offsets based on explicit signals are introduced to capture facial variations on diverse avatar styles. We also inject explicit hand controls for more accurate and realistic hand textures and movements. Additionally, a facial refinement module is employed to enhance facial realism ensuring highly expressive and lifelike portrait videos.
    \item \textbf{Extensible and real-time generation framework} is constructed for interactive video chat application, which can adapt to various scenarios through flexible sub-module combinations, supporting tasks ranging from head-driven animation to upper-body generation with hand gestures. Besides, we establish an efficient streaming inference pipeline that achieves 30fps at a resolution of maximum $512 \times 768$ on 4090 GPU, ensuring smooth and immersive experiences in real-time video chat.
\end{itemize}

Our experimental results highlight the effectiveness of our proposed approach in generating portrait videos that exhibit both rich facial expressiveness and upper-body movements. This capability is particularly valuable for creating engaging and lifelike digital human interactions.

\section{Related Work}
\label{sec:relatedwork}

\begin{figure*}[t]
  \centering
   \includegraphics[width=0.9\linewidth]{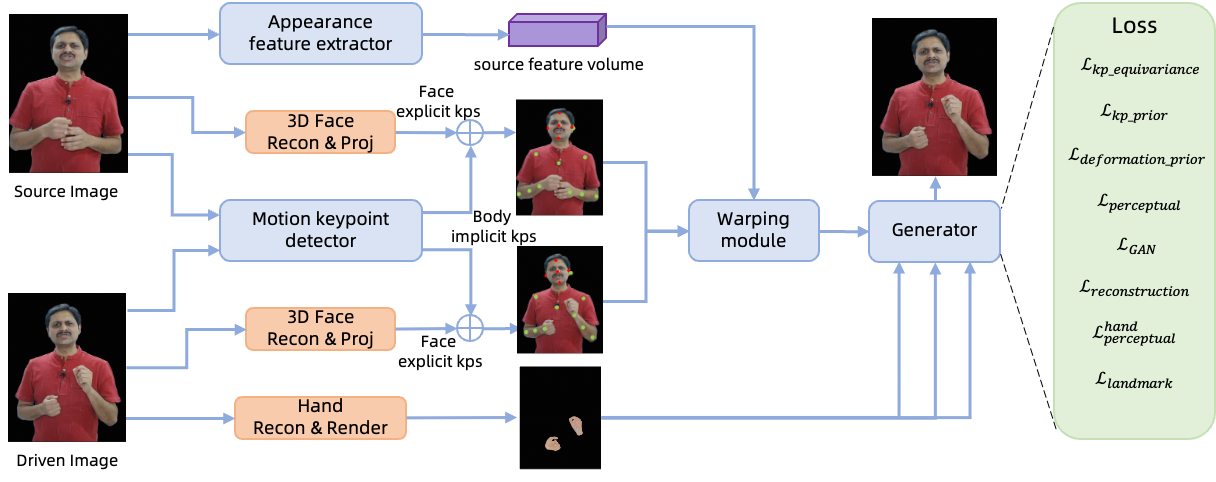}

   \caption{Pipeline of upper-body video generation with hybrid control fusion, which takes both explicit facial keypoints and implicit body keypoints to conduct feature warping, while rendered hand image further inject into generator for improving the quality of hand generation.}
   \label{fig:videodriven}
\end{figure*}

In this section, we focus on reviewing existing methods that utilize a two-stage framework for audio-driven portrait video synthesis. The first stage involves generating motion representations from audio inputs, while the second stage aims at transforming motion representation into high-quality digital human videos.

\subsection{Audio to motion representation}

The first stage converts audio inputs into detailed facial motion representations, and various techniques have been proposed to achieve this. Some early methods \cite{hdtf, pirender, sadtalker} take the parameters of three-dimensional morphable face models (3DMMs) to represent facial motions. In these methods, identity, facial expressions and head pose are decomposed into separate coefficients, supporting independent control over each aspect of the face. However, 3DMM-based methods have limitations in capturing fine-grained facial details due to the parameterized facial model. 
Subsequent works \cite{makeittalk, aniportrait} introduce explicit facial landmark or mesh as control signals, which offer more flexible and expressive representation, and have strong generalization capabilities, enabling the facial control of non-human characters. Despite these improvements, landmark-based methods still face challenges in capturing subtle variations and nuances in facial expressions. 

The latest advancements propose the use of implicit control signals, such as implicit keypoints or motion latent variables to achieve better facial expression control. EAT \cite{emotalk} and Ditto \cite{ditto} employ 3D implicit keypoints for facial motion control and further apply emotion label as additional condition. While AniTalker \cite{anitalker} and VASA-1 \cite{vasa} extract implicit motion latent variables from video sequences to capture more detailed movements, and construct diffusion motion generator to perform emotionally rich facial expressions. However, these fully implicit motion representations still have the challenge of completely decoupling identity and expression, leading to potential artifacts where identity might inadvertently change along with expression.

To address these limitations, we explore hybrid multimodal model to combine explicit and implicit motion representations for fully utilizing their complementary advantages. Beyond facial expressions, some recent works \cite{latentCospeech,vlogger, emo2} have extended their focus to predict control signals for upper-body movements, though further exploration is still needed to develop effective coordination between facial and body movements.

\subsection{Portrait video generation}
The second stage focuses on the generation of high-quality digital human video frames, conditioned on the motion representations predicted from the first stage by audio inputs.
Early works \cite{makeittalk,livespeech} in this domain primarily rely on Generative Adversarial Networks (GANs) \cite{gan} for generating portrait video frames. Furthermore, some researchers explore warp-based methods \cite{emotalk, anitalker, vasa, latentCospeech} that learn dense motion flow from source image to target, achieving better results in terms of preserving image quality and detail. Despite their generative power, GANs still struggle with high-quality detail generation, especially in facial expressions and body movements as well as their complex interactions.

With the advent of neural rendering techniques, some methods adopt Neural Radiance Fields (NeRF) \cite{adnerf,sspnerf,geneface} or Gaussian Splatting \cite{gaussiantalker} for portrait video generation, which offer superior rendering quality. These neural rendering based methods are capable of generating highly photorealistic portrait videos, supporting real-time generation at the same time. However, a notable drawback of these methods is the need for retraining on each specific character, which limits their scalability.

More recently, diffusion models have shown remarkable progress in generating high-quality digital human videos. Methods like AniPortrait \cite{aniportrait}, Vlogger \cite{vlogger}, EMO2 \cite{emo2} inject face or body control signals into denoising UNet along with reference net to generate talking portrait video. However, diffusion methods face challenges in generating long-duration videos stably and efficiently. Their inference process is computationally intensive and slow, making real-time generation difficult. 
Given the strengths and weaknesses of the above approaches, we opt to employ GANs as our primary generator aiming to balance the inference efficiency and performance for both facial and upper-body generation.

\section{Method}
\label{sec:method}

Our proposed approach generates high-quality, stylized portrait videos driven by audio inputs. It takes a character image (either a head shot or upper-body shot) and an audio clip as primary inputs, with an optional reference video for style transfer.
The process consists of two main stages: first, we convert audio input into motion representations that capture facial expressions and body movements; second, we synthesize the portrait video using these motion representations and the character image. If provided, the reference video guides the style transfer to ensure the generated video reflects the desired stylistic elements.

In the following subsections, we first give preliminaries of our approach, then motion representation learning is introduced. Next we present the module that generates motion representations from audio inputs. Finally, we describe the video generation based on motion representations.

\subsection{Preliminaries}

To balance the quality of generated videos and model inference performance, we employ a warping-based GAN framework inspired by existing works \cite{facevid2vid, liveportrait}. The general framework consists of four key components as shown in Figure \ref{fig:videodriven}: 1) \textit{Appearance feature extraction} to get visual features from source image; 2) \textit{Motion representation extraction} that extracts motion to represent facial expressions and body movements, described in Section \ref{mr}; 3) \textit{Warping field estimation} to calculate the transformation from source to target; and 4) \textit{Generator} to synthesize final images from warped appearance features, introduced in Section \ref{vg}. 

To generate high expressive motion representations from audio inputs, our approach is based on the diffusion model \cite{ddpm}. In the forward process, noise $\epsilon \sim N(0,1)$ is gradually added to the original data $x_0$ over a series of time steps $t$. The reverse process aims to remove the noise and recover the original data. This is achieved using a learned neural network $\epsilon_\theta(x_t,c,t)$, where $c$ indicates that control signals in our approach discussed in Section \ref{a2m}, {\em e.g.} audio and reference motions. The training objective is defined as follows:
\begin{equation}
    \begin{aligned}
        \mathcal{L}=\mathbb{E}_{t,c,x_t,\epsilon }\left [ \left \| \epsilon - \epsilon_\theta (x_t,t,c)  \right \| ^2 \right ] 
    \end{aligned}
\end{equation}
which minimizes the difference between predicted noise $\epsilon_\theta(x_t,c,t)$ and actual noise.

\subsection{Motion representation}
\label{mr}
To balance the expressiveness and controllability of face and upper-body movements, we consider three types of motion representations commonly used in existing methods. Explicit keypoints \cite{AA} and body templates \cite{graph} provide clear and editable control over movements, while implicit keypoints \cite{liveportrait} offer richer detail information. To leverage the strengths of all three representations, we adopt a hybrid representation that combines their advantages.
Specifically, head motion is represented by 3D keypoints projected from a 3DMM head template \cite{faceverse}, upper body motion is derived from implicit keypoints, while hand motion is captured using rendered images based on the MANO template \cite{mano}. %

For head motion, 3D keypoints $X_{ori}^{head}$ on key face regions are projected from the head template, including eyes, mouth, eyebrows and face contours, driven by 3DMM coefficients, which are capable of representing a diverse range of facial expressions. However, restricted by the head template's expressive capabilities, these keypoints cannot represent extreme facial features, such as large eyes or twisted mouth. To address this issue, we propose a keypoints displacement module that aligns the keypoints with actual positions of facial features from image by adding an implicit offset $\Delta X^{head}$ to the original explicit keypoints $X_{ori}^{head}$, as shown in the right of Figure \ref{fig:faceoffset}, which can capture more detailed facial expressions. The final head keypoints $X^{head}$ are represented as:
\begin{equation}
    \begin{aligned}
        X^{head} = X_{ori}^{head} + \Delta X^{head}
    \end{aligned}
    \label{eq:offset}
\end{equation}

For upper body motion, we adopt 3D implict keypoints similar to LivePortrait \cite{liveportrait}. Specifically, given the canonical keypoints $X_{c}^{body} \in \mathbb{R}^{k \times 3}$ extracted from the image, along with body movement $\delta \in \mathbb{R}^{k \times 3}$, rotation $R \in \mathbb{R}^{3 \times 3}$, scale factors $s \in \mathbb{R}^{1}$, and translations $t \in \mathbb{R}^{3}$. The rotation $R$ is set as the identity matrix that simplifies body movements into planar motion. The final driving 3D implicit keypoints $X^{body}$ can be represented as:
\begin{equation}
    \begin{aligned}
        X^{body} = s(X_{c}^{body}R + \delta) + t
    \end{aligned}
    \label{eq:body}
\end{equation}

For hand motion, we adopt hand coefficients to render hand images from MANO template as control signals, using textures derived from our collected high-resolution data. The pixel-aligned hand control signals are particularly beneficial for preserving the hand shape and ensuring high-quality hand gesture generation.

\begin{figure}[t]
  \centering
   \includegraphics[width=1.0\linewidth]{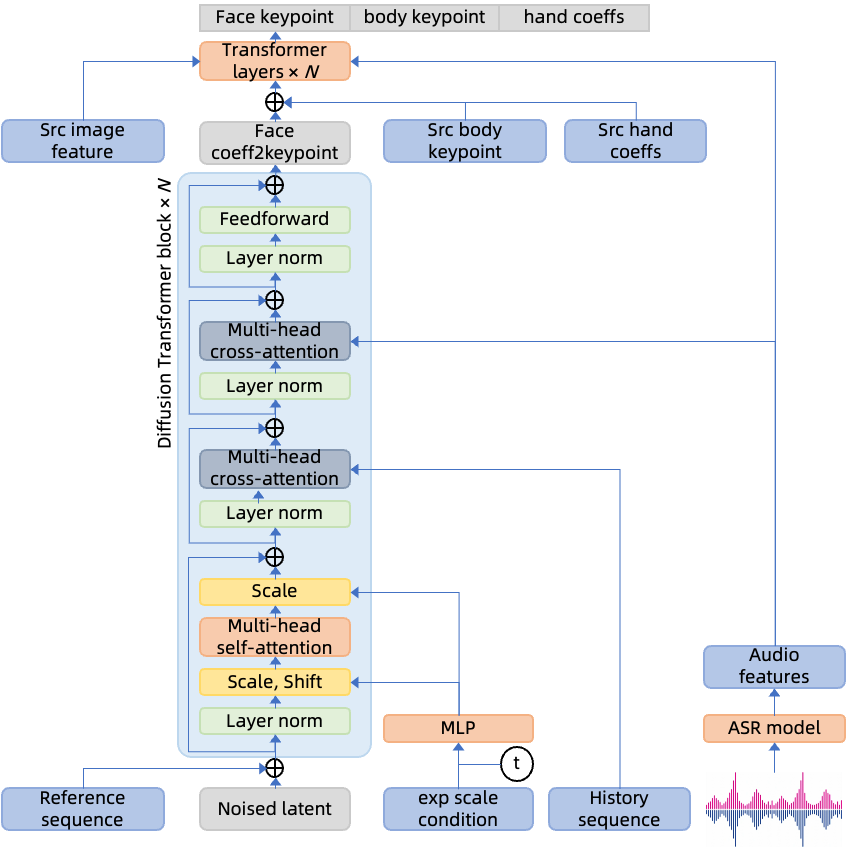}

   \caption{Illustration of hierarchical audio2motion diffusion model, including facial motion prediction with style control at bottom, and upper-body motion prediction with hands at top.}
   \label{fig:a2m}
\end{figure}

\subsection{Audio to motion representation}
\label{a2m}

To effectively generate motion representations from audio inputs, especially in the context of upper-body driving, we design a hierarchical model, which is inspired by the observation that human upper-body movements are often driven by head motions, while hand movements mainly follow the rhythm of audio signal. Thus, as shown in Figure \ref{fig:a2m}, our hierarchical audio driving model first predicts facial motion from audio inputs, and then generates upper-body (including hands) motions driven by the previously predicted facial motion.

\noindent \textbf{Facial motion prediction with style control}.
In the first sub-stage, facial motion is predicted directly from the audio inputs. Specifically, we use the expression blendshapes and 6-dimensional pose coefficients including rotation and translation defined on the head template \cite{faceverse} as facial motion representation, with the goal of simple, effective and efficient model learning. For the input audio clip, we extract audio features from a transformer-based speech recognition model \cite{paraformer}, containing rich semantic information that guides the generation of lip movements and facial expressions.

The core face driving model is based on diffusion transformer \cite{dit}. We inject audio features into the network using cross-attention mechanisms. To ensure temporal coherence, we also incorporate historical motion information via cross-attention for smooth transitions between consecutive frames. Besides, for style control, we use Adaptive Layer Normalization (AdaLN) to inject coarse-grained expression and pose range information into the network, enabling precise control over the amplitude of facial movements. Additionally, a facial expression sequence extracted from reference video by 3D face reconstruction, can be optionally provided for style transfer. The reference sequence is concatenated along the temporal dimension with the noise latent, which can effectively transfer the style from the reference video to the final output, enabling high controllable facial expression generation.

\noindent \textbf{Upper-body motion prediction with hands}.
In the second sub-stage, we extend the facial motion predictions to drive upper-body movements. Considering the discrepancies between facial coefficients and body keypoints, as illustrated in the top of Figure \ref{fig:a2m}, we first project the facial parameters into 3D keypoints through head template. These projected keypoints are served as conditions injected into transformer network. Simultaneously, the aforementioned audio features as well as appearance feature from source image are also injected through cross-attention.

The model predicts facial keypoint offsets $\Delta X^{head}$, as mentioned in Section \ref{mr} to refine facial expression details and generates upper-body keypoints along with hand control coefficients. This ensures that the upper-body motions naturally follow the head's dynamics while incorporating hand gestures synchronized with the audio rhythm, which can generate natural and coherent body motions as well as expressive and realistic gestures that match the spoken content for lifelike portrait video.

\subsection{Video generation}
\label{vg}
We employ a warping-based GAN framework similar to \cite{liveportrait} for generating upper-body image from motion representation. The entire process is divided into two stages: upper body image generation and face refinement, which share a similar network architecture.

\noindent \textbf{Upper body image generation}.
The framework shown in Figure \ref{fig:videodriven} includes an appearance feature extractor $\mathcal{F}$, a keypoint detector $\mathcal{L}$, a warping field estimator $\mathcal{W}$ and a generator $\mathcal{G}$. Given a source image, $\mathcal{L}$ detects canonical body keypoints $X_{c,s}^{body}$, body movement $\delta$ and translation $t$. The source 3D implicit body keypoints $X_{s}^{body}$ can be obtained according to Eq. \ref{eq:body}. Notably, we select explicit 3D face keypoints $X_{s}^{head}$ from the head template as additional control signals, and incorporate them with implicit body keypoints. Ultimately, the control signals received by $\mathcal{W}$ can be represented as:
\begin{equation}
    \begin{aligned}
        X_{s}^{full} = X_{s}^{head} \oplus X_{s}^{body}
    \end{aligned}
    \label{eq:full}
\end{equation}

The source feature volume derived from $\mathcal{F}$ is warped to the target feature volume based on source $X_{s}^{full}$ and driving $X_{d}^{full}$ through $\mathcal{W}$. Subsequently, a generator $\mathcal{G}$ processes the warped features and generates target image. Besides, we also train a face animation version with the same architecture driving from $X_s^{head}$ to $X_d^{head}$, realizing talking head generation.

\noindent \textbf{High quality hand generation}.
The structure of hand is complex and prone to self-occlusion, making it challenging to generate high-quality hand images based solely on 3D implicit keypoints. 
To address this issue, we first utilized Image Quality Assessment (IQA) \cite{pyiqa} to filter out low-quality hand data. Then, we employ the MANO template to render hand images and inject them into the generator, thereby providing stronger prior information to facilitate hand generation. Specifically, given that a certain layer of the generator has feature values denoted as $f$, we introduce an additional AdaIN module to incorporate the supplementary information from rendered hand image $I_{hand}$, which is represented as:
\begin{equation}
    \begin{aligned}
        f = f + AdaIN(I_{hand})
    \end{aligned}
    \label{eq:adain}
\end{equation}
The rendered hand image $I_{hand}$ is injected across multiple scales within the generator, significantly improving the quality of hand generation with negligible additional computational cost.

\noindent \textbf{Cascaded loss terms}. 
We follow LivePortrait \cite{liveportrait} to use implicit keypoints equivariance loss $\mathcal{L}_{E}$, keypoint prior loss $\mathcal{L}_{L}$, deformation prior loss $\mathcal{L}_{D}$, perceptual loss $\mathcal{L}_{Per}$, GAN loss $\mathcal{L}_{GAN}$, reconstruction loss $\mathcal{L}_{Recon}$. To enhance the quality of hand region, we obtain the hand mask using the MANO model and introduce a hand region perceptual loss $\mathcal{L}_{Per}^{Hand}$. For more precise representation of limb movements through implicit keypoints, we adopt 2D landmark loss $\mathcal{L}_{lms}$ to optimize the positions of implicit keypoints similar to \cite{liveportrait}. The overall training objective is formulated as follows:
\begin{equation}
    \begin{aligned}
        \mathcal{L} &= \mathcal{L}_{E}+\mathcal{L}_{L}+\mathcal{L}_{D}+\mathcal{L}_{Per} \\
        &+\mathcal{L}_{GAN}+\mathcal{L}_{Recon}+\mathcal{L}_{lms}+\mathcal{L}_{Per}^{Hand}
    \end{aligned}
    \label{eq:loss}
\end{equation}

\begin{figure}[t]
  \centering
   \includegraphics[width=1.0\linewidth]{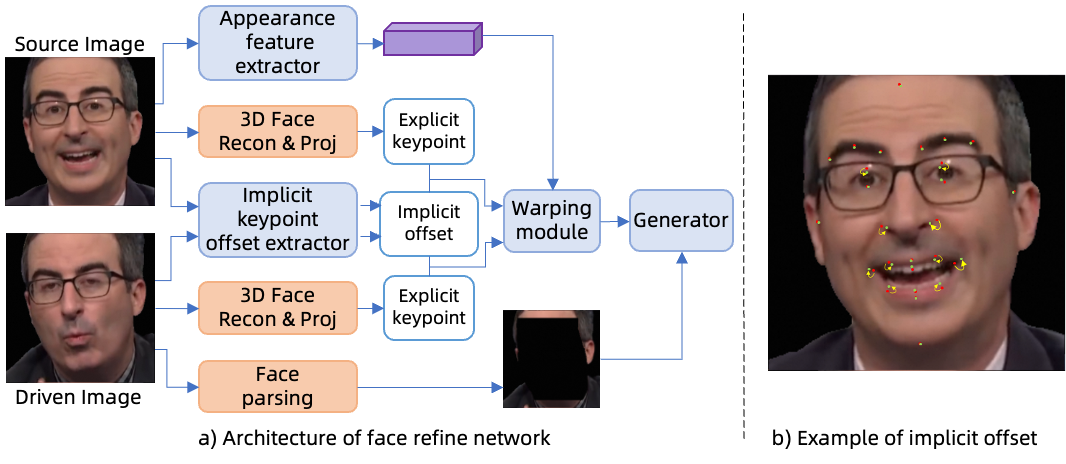}

   \caption{Illustration of face refine network, the left of figure shows the architecture, while the right demonstrates that more precise facial keypoints are located by adding implicit offset.}
   \label{fig:faceoffset}
\end{figure}

\noindent \textbf{Face refine network}. 
Restricted by the generative capabilities of GANs and the relatively small proportion of the face in images, the aforementioned network fails to produce satisfactory facial details, which significantly diminishes overall realism. To overcome this, we introduce a lightweight face refine network, driven by 3D explicit head keypoints with implicit offset as described in Equation \ref{eq:offset}, as shown in Figure \ref{fig:faceoffset}, which shares a similar structure with the body generation network but with the following distinctions. 
Specifically, the input for the generator's AdaIN module is the cropped background image of the head, where the facial region is masked out, ensuring that the restored face can seamlessly blend with the generated half-body image.
The training loss is similar to that used in the upper-body training, with hand region perceptual loss $\mathcal{L}_{Per}^{Hand}$ replaced by facial region loss $\mathcal{L}_{Per}^{face}$. Additionally, we omit the 2D landmark loss $\mathcal{L}_{lms}$.

Finally, our generative network produces upper-body portrait images with hand motion, and the facial region is specially refined and reintegrated into the original image to obtain the final result. This strategy ensures that the generated image retains rich details of facial expressions while maintaining the visual quality of body movements.

\begin{table*}
  \centering
  \begin{tabular}{@{}lccccccc|c@{}}
    \toprule
    Method & PSNR$\uparrow$ & SSIM$\uparrow$ & LPIPS$\downarrow$ & FID$\downarrow$ & FVD$\downarrow$  & CSIM$\uparrow$ & HKC$\uparrow$ & FPS (Resolution) \\
    \midrule
    FOMM \cite{fomm} & 18.92 & 0.677 & 0.269 & 42.690 & 569.893 &  0.525 & 0.494 & 87 (256*256) \\
    MRAA \cite{mraa} & 19.12 & 0.696 & 0.253 & 35.546 & 419.293 &  0.536 & 0.534 & 77 (384*384) \\
    LIA \cite{lia} & 18.96 & 0.681 & 0.258 & 44.747 & 387.924 &  0.590 & 0.548 & 30 (256*256) \\
    TPSMM \cite{tpsmm} & 19.64 & 0.707 & 0.237 & 34.509 & 384.663 &  0.597 & 0.567 & 48 (384*384) \\
    \midrule
    w/o hand injection & 24.59 & 0.829 & 0.132 & 6.825 & 38.401  & 0.605 & 0.607 & 34 (512*768) \\
    w/o face refine & 24.87 & 0.829 & 0.126 & 5.799 & 34.124  & 0.613 & 0.652 & 37 (512*768) \\
    \textbf{Ours} & \textbf{24.88} & \textbf{0.831} &\textbf{0.126} & \textbf{5.505} & \textbf{33.349}  & \textbf{0.654} & \textbf{0.652} & 33 (512*768)\\
    \midrule
    w/o facial hybrid control* & 22.85 & 0.799 & 0.170 & 6.355 & 64.249  & 0.627 & - & 40 (512*512)\\
    \textbf{Ours*} & \textbf{23.09} & \textbf{0.807} & \textbf{0.166} & \textbf{6.297} & \textbf{47.914}  & \textbf{0.632} & - & 41 (512*512) \\
    \bottomrule
  \end{tabular}
  \caption{Quantitative comparisons of upper-body video generation under self-driven reenactment mode. The inference speed, measured in frames per second (FPS), and the resolution of the generated output are also presented in the table. The * symbol denotes that the evaluations are conducted on talking head video reenactment to verify the effectiveness of implicit facial keypoint offset.}
  \label{tab:videodriven}
\end{table*}

\section{Experiments}

In this section, we introduce our experimental setup and results. We begin with the implementation details, followed by presenting results under both video-driven and audio-driven setup compared with existing methods. Finally, we conduct ablation studies to validate the effectiveness of different modules in our approach.

\subsection{Implementation details}

We independently train two-stage models for audio-driven and video generation tasks. For audio to motion representation part, we use a 6-layer diffusion transformer to predict facial coefficients, while the following upper-body motion prediction model adopts a 2-layer transformer. Both modules are trained on an A100 GPU with a batch size of 16 for 100K steps. For video generation part, we have different resolutions depending on whether we generate upper-body or head-only images. Upper-body generation outputs at 512 $\times$ 768, while head-only generation has 512 $\times$ 512 output resolution. All models utilize Adam optimizer with a learning rate of 1e-4, trained on 8 A100 GPUs for 7 days with a batch size of 32. During inference, the video generation modules run serially and achieve a real-time inference speed of 30fps on 4090 GPU. More details of our implementation as well as inference pipeline are presented in supplementary materials.

For training and testing data, we collect and clean about 20,000 talkshow video clips from YouTube, totaling approximately 30 hours at 30 fps, and we randomly select 500 clips for testing during comparison with existing methods as well as our ablation studies. 

\begin{table}
  \centering
  \begin{tabular}{@{}lcc@{}}
    \toprule
    Method & HKC$\uparrow$ & CSIM$\uparrow$ \\
    \midrule
    CyberHost* \cite{cyberhost} & \textbf{0.723} & \textbf{0.706}  \\
    \textbf{Ours*} & 0.708 & 0.657 \\
    \midrule
    EchoMimicV2* \cite{echomimicv2} & 0.618 & 0.621  \\
    \textbf{Ours*} & \textbf{0.642} & \textbf{0.683} \\
    
    \bottomrule
  \end{tabular}
  \caption{Quantitative comparisons of video generation with diffusion based method. We reenact the demo videos provided by the compared methods and evaluate the results of hand generation. }
  \label{tab:diffusion}
\end{table}

\subsection{Video generation results}

Given the limited number of GAN-based upper-body audio-driven works, we first independently validate the effectiveness of our video generation model in producing upper-body videos through a self-driven reenactment setup. This validation is crucial as it directly determines the quality of the generated portrait video in upper-body audio-driven scenarios. 
We compare our approach against several existing GAN-based video-driven methods, including FOMM \cite{fomm}, MRAA \cite{mraa}, LIA \cite{lia}, and TPSMM \cite{tpsmm}, using multiple metrics to assess the quality of the generated frames: Peak Signal-to-Noise Ratio (PSNR), Structural Similarity Index (SSIM) \cite{ssim}, Learned Perceptual Image Patch Similarity (LPIPS) \cite{lpips}, Fr\'echet Inception Distance (FID) \cite{fid}, and Fr\'echet Video Distance (FVD) \cite{fvd}. We also compute the cosine similarity (CSIM) between the features of reference image and generated video frames. Additionally, we calculate Hand Keypoint Confidence (HKC) to evaluate the quality of hand representation in generated frames.

\begin{figure*}[t]
  \centering
   \includegraphics[width=1.0\linewidth]{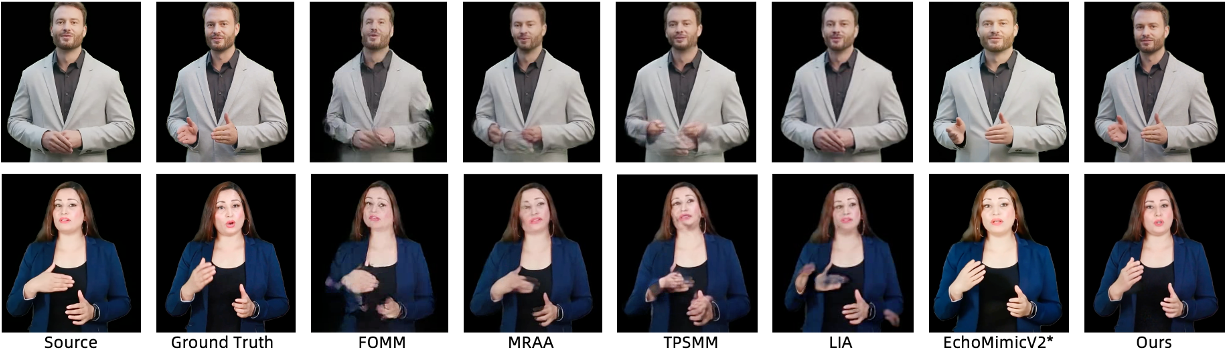}

   \caption{Qualitative comparisons of upper-body video generation under self-driven reenactment setup. Our approach significantly outperforms the GAN-based comparison methods, and achieves comparable quality with the diffusion-based method EchoMimicV2.}
   \label{fig:vis_upperbody}
\end{figure*}

We present quantitative comparison results in Table \ref{tab:videodriven}, and we also provide qualitative comparisons through visual examples as shown in Figure \ref{fig:vis_upperbody}. The results clearly show that our proposed approach significantly outperforms existing methods in terms of upper-body generation quality, achieved by hybrid control fusion of both explicit and implicit motion representation for video generation.
Specifically, our approach demonstrates superior performance in generating hand movements, as evidenced by qualitative comparisons where the quality of hand generation is notable better than that of existing methods, which is thanks to our proposed high-quality hand generation by incorporating the rendered hand images as supplementary information. 

Besides, we further compared with latest diffusion based methods as shown in Table \ref{tab:diffusion}, including CyberHost \cite{cyberhost} and EchoMimicV2 \cite{echomimicv2} to evaluate the upper-body generation with hand, and we can observe that we still achieve comparable results that indicate the effectiveness of our approach.
Additionally, the inference speeds are also presented in Table \ref{tab:videodriven}. Although the compared methods achieve higher FPS, their model output resolutions are relatively low. In contrast, our approach achieves real-time inference at 30+FPS with a higher output resolution as well as superior quantitative performance, which highlights the advantages of our solution, balancing both quality and efficiency, and making it suitable for practical applications.

\begin{table}
  \centering
  \begin{tabular}{@{}lcccc@{}}
    \toprule
    Method & FID$\downarrow$ & CSIM$\uparrow$ & Sync$\uparrow$ & Diversity$\uparrow$ \\
    \midrule
    SadTalker \cite{sadtalker} & 52.32 & 0.595 & 4.120 & 0.112 \\
    AniTalker \cite{anitalker} & 19.74 & 0.578 & 4.066 & 0.099 \\
    \textbf{Ours} & \textbf{9.49} & \textbf{0.668} & \textbf{5.668} & \textbf{0.137}\\
    \bottomrule
  \end{tabular}
  \caption{Quantitative comparisons of audio-driven results under head-only talking animation generation.}
  \label{tab:audiodriven}
\end{table}

\subsection{Audio driven results}

In this subsection, we focus on comparing the results of audio-driven portrait videos, with an emphasis on lip-sync accuracy and motion diversity. Given the limited number of GAN-based works directly addressing audio-driven upper-body motion generation, we conduct a targeted comparison with existing GAN-based talking head methods, specifically evaluating the quality of facial generation under audio-driven conditions. The primary compared methods include SadTalker \cite{sadtalker} ad AniTalker \cite{anitalker}. Our evaluation metrics include not only the image quality metrics mentioned earlier, but also the Sync metric, proposed by SyncNet \cite{syncnet} to assess the synchronization between lip movements and audio signals, and face keypoint variance indicating the diversity of facial movements.

We present the quantitative experimental results in Table \ref{tab:audiodriven}, while visual comparisons of talking head generation results from different methods are provided in supplementary materials. These quantitative comparisons and visual examples highlight the differences in lip-sync accuracy and motion diversity among the compared methods. The results demonstrate that our proposed approach outperforms existing methods, indicating better alignment between mouth movements and corresponding audio signals and more varied facial expressions. These enhancements can be attributed to our hierarchical motion diffusion model, which combines explicit and implicit control signals for more precise and expressive facial and limb movements.

\subsection{Ablation studies}

To evaluate the effectiveness of various sub-modules in our proposed approach, we conduct comprehensive ablation studies, including facial style control and transfer capability, the impact of explicit and implicit hybrid control fusion, the effect of hand signal injection, and the performance of face refinement. Visual comparisons of these ablation studies are shown in supplementary materials.

\begin{table}
  \centering
  \begin{tabular}{@{}lcc@{}}
    \toprule
    Method & MAE$\downarrow$ & SSIM$\uparrow$  \\
    \midrule
    w/o style transfer & 0.074 & 0.373 \\
    \textbf{Ours} & \textbf{0.049} & \textbf{0.709} \\
    
    \bottomrule
  \end{tabular}
  \caption{Comparison on reference style transfer, calculated on the face coefficients predicted from audio2motion model.}
  \label{tab:style}
\end{table}

\noindent \textbf{Facial style control and transfer}. 
We evaluate the ability to control and transfer facial expression styles. First, we compare the impact of injecting expression motion sequences from reference video on the final generated facial coefficients. The similarity between the generated facial coefficients and the ground-truth is measured by MAE and SSIM. The results presented in Table \ref{tab:style} show that injecting expression information from reference video can improve the similarity to ground-truth, indicating that the expression style in reference video is effectively transferred into the generated results. 
Besides, qualitative comparisons shown in supplementary materials further validate the influence of explicitly controlling the magnitude of expressions. 

\noindent \textbf{Explicit and implicit hybrid control fusion}.
We examine the effect of combining explicit and implicit keypoints on facial expression image generation. Specifically, we compare the results with and without the implicit keypoint offsets, as described in Equation \ref{eq:offset}. The comparison results, shown in Table \ref{tab:videodriven}, indicate that incorporating implicit keypoint offsets improves the quality and accuracy of generated facial expression images. This enhancement suggests that introducing hybrid control helps to increase the flexibility and diversity of audio-driven facial expressions, while maintaining high controllability at the same time.

\noindent \textbf{Hand signal injection}.
We also investigate the impact of injecting hand control signals on the quality of upper-body image generation. The results in Table \ref{tab:videodriven} reveal a significant degradation in image quality when hand control signals are not injected. This finding indicates the critical role of hand signal injection, enabling the network to produce accurate and clear hand movements. The addition of hand control signals ensures that the generated upper-body images are more realistic and coherent.

\noindent \textbf{Face refinement}.
We finally assess the effectiveness of the face refinement module. The results are presented in Table \ref{tab:videodriven}, which show a notable improvement in the quality of generated facial images when face refine is applied. This indicates that face refine enhances the overall realism and expressiveness of the portrait video by improving the quality of facial region. The enhanced facial details contribute to a more lifelike upper-body portrait video.

\section{Conclusion}

In this paper, we introduce a novel stylized portrait video generation approach using a hierarchical motion diffusion audio-driven model and a hybrid control fusion video generation model. Our two-stage extensible framework supports tasks from head-driven animations to upper-body videos with hand movements, achieving expressive lifelike portrait video generation, and ensuring inference at 30fps on a 4090 GPU, which can provide a robust solution for real-time and high-quality video-chat interactions with various applications such as virtual avatars, live streaming, and augmented reality.

{
    \small
    \bibliographystyle{ieeenat_fullname}
    \bibliography{main}
}

\end{document}